%% file: main.tex
\documentclass[10pt,twocolumn,letterpaper]{article}

\usepackage{cvpr}              


\usepackage{amsmath,amssymb,amsfonts}
\usepackage{graphicx}
\usepackage{color}
\usepackage[table]{xcolor}
\usepackage{censor}
\usepackage{fix-cm}
\usepackage{textcomp}
\usepackage{tikz}
\usetikzlibrary{calc,decorations,decorations.markings,decorations.text}
\usepackage{bm}
\usepackage{colortbl}
\usepackage{siunitx}
\usepackage[nolist,nohyperlinks]{acronym}
\usepackage{subcaption}
\usepackage{gensymb}
\usepackage{booktabs}
\usepackage{caption}
\usepackage{multirow}
\usepackage{algpseudocode}
\usepackage{algorithm}
\def\BibTeX{{\rm B\kern-.05em{\sc i\kern-.025em b}\kern-.08em
    T\kern-.1667em\lower.7ex\hbox{E}\kern-.125emX}}
\AtBeginDocument{\definecolor{tmlcncolor}{cmyk}{0.93,0.59,0.15,0.02}\definecolor{NavyBlue}{RGB}{0,86,125}}

\definecolor{cvprblue}{rgb}{0.21,0.49,0.74}
\usepackage[pagebackref,breaklinks,colorlinks,allcolors=cvprblue]{hyperref}
\usepackage[capitalize]{cleveref}
\crefname{section}{Sec.}{Secs.}
\Crefname{section}{Section}{Sections}
\Crefname{table}{Table}{Tables}
\crefname{table}{Tab.}{Tabs.}

\acrodef{POV}[POV]{Point Of View}
\acrodef{ROI}[ROI]{Region Of Interest}
\acrodef{DLT}[DLT]{Direct Linear Transform}
\acrodef{FOV}[FOV]{Field Of View}
\acrodef{PCA}[PCA]{Principal Component Analysis}
\acrodef{COR}[COR]{Coefficient Of Restitution}
\acrodef{PnP}[PnP]{Perspective-n-Point}
\acrodef{ODE}[ODE]{Ordinary Differential Equation}
\acrodef{MAE}[MAE]{Mean Absolute Error}
\acrodef{MRAE}[MRAE]{Mean Relative Absolute Error}
\acrodef{mIoU}[mIoU]{mean Intersection over Union}
\acrodef{GRU}[GRU]{Gated Recurrent Unit}
\acrodef{MLP}[MLP]{Multi-Layer Perceptron}
\acrodef{SNR}[SNR]{Signal Noise Ratio}
\acrodef{SE}[SE]{Squeeze-and-Excitation}
\acrodef{ECA}[ECA]{Efficient Channel Attention}
\acrodef{CA}[CA]{Coordinate Attention}
\acrodef{CBAM}[CBAM]{Convolutional Block Attention Module}
\acrodef{GC}[GC]{Global Context}
\acrodef{RMSE}[RMSE]{Root Mean Square Error}
\acrodef{ViT}[ViT]{Vision Transformer}

\def\paperID{*****} 
\def\confName{CVPR}
\def\confYear{2026}

\title{BlurBall: Joint Ball and Motion Blur Estimation for Table Tennis Ball Tracking}

\author{Thomas Gossard \quad Filip Radovic \quad Andreas Ziegler \quad Andreas Zell
\thanks{This research was partially funded by Sony AI.}\\
University of Tuebingen\\
{\tt\small \{thomas.gossard,filip.radovic,andreas.ziegler,andreas.zell\}@uni-tuebingen.de}
}

\begin{document}
\maketitle
\input{sec/0_abstract}    
\input{sec/1_intro}
\input{sec/2_related_work}
\input{sec/3_method}
\input{sec/4_experiments}
\input{sec/5_conclusion}

{
    \small
    \bibliographystyle{ieeenat_fullname}
    \bibliography{main}
}


\input{sec/appendix}

\end{document}

%% file: sec/0_abstract.tex
\begin{abstract}
Motion blur reduces the clarity of fast-moving objects, posing challenges for detection systems, especially in racket sports, where balls often appear as streaks rather than distinct points. 
Existing labeling conventions mark the ball at the leading edge of the blur, introducing asymmetry and ignoring valuable motion cues correlated with velocity.
This paper introduces a new labeling strategy that places the ball at the center of the blur streak and explicitly annotates blur attributes.
Using this convention, we release a new table tennis ball detection dataset.
We demonstrate that this labeling approach consistently enhances detection performance across various models.
Furthermore, we introduce BlurBall, a model that jointly estimates ball position and motion blur attributes.
By incorporating attention mechanisms such as Squeeze-and-Excitation over multi-frame inputs, we achieve state-of-the-art results in ball detection. 
Leveraging blur not only improves detection accuracy but also enables more reliable trajectory prediction, benefiting real-time sports analytics.
Project page: \url{https://cogsys-tuebingen.github.io/blurball/}
\end{abstract}

%% file: sec/1_intro.tex
\section{Introduction}
\label{sec:intro}

\begin{figure}
    \centering
    \includegraphics[width=0.9\linewidth]{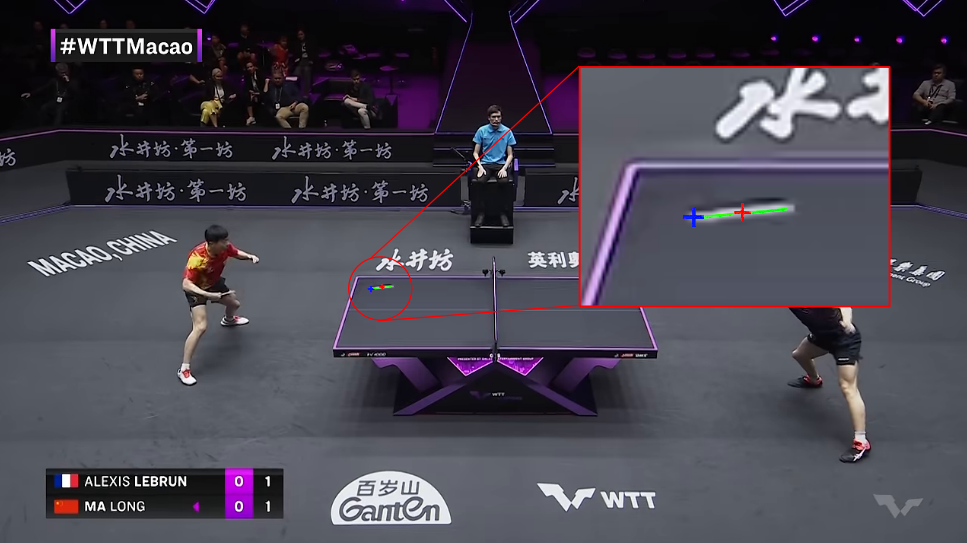}
    \caption{Motion blur frequently appears in broadcast footage but is typically disregarded. Yet, it offers valuable cues for estimating the ball’s velocity. The blue cross denotes the classical labeling approach, which introduces asymmetry and ambiguity to the detection task. We propose a refined annotation strategy: relabel the ball center to the middle of the blur (red cross) and include a directional blur label (green line) to better capture motion information.}
    \label{fig:blur_example}
\end{figure}

Sports analytics has become a vital component in understanding athletic performance, refining strategies, and enhancing athlete evaluation.
Extracting meaningful insights from raw game footage has traditionally been both time-consuming and costly.
Nevertheless, the benefits, ranging from improved performance metrics to deeper tactical understanding, are substantial.
Recent advances in computer vision now enable automatic extraction of key game elements such as ball trajectories, player poses, and even racket orientation~\cite{naik2022}.
These capabilities not only support retrospective analysis but also enable predictive applications, for example, forecasting ball bounce positions from serve strokes~\cite{wu2019}, or identifying the characteristics of a winning rally~\cite{muelling2014c}.
Such techniques also open the door to automated umpiring systems based solely on video footage~\cite{wong2023}.

Moreover, this data is particularly promising for training table tennis robots.
Despite remarkable progress~\cite{dambrosio2024, tebbe2019, kober2011, Kyohei2019ThePP}, incorporating human-like behavior remains challenging due to the lack of diverse, high-quality datasets.
Ideally, the abundance of online match recordings could be leveraged to train robust, human-informed policies.
For example, \cite{etaat2025} introduced a 20-hour dataset collected from real matches and demonstrated its utility in simulation, thereby improving ball-return prediction and enhancing robotic performance.

These applications rely on accurate and robust ball tracking, an essential prerequisite for intelligent analysis, forecasting, and autonomous behavior in high-speed sports environments.
However, achieving this is far from trivial, especially in racket sports like table tennis, where the ball generally moves at extreme speeds.
The ball's high speed often results in motion blur, transforming its appearance from a distinct dot into a streak of varying length.
This motion blur not only complicates detection but also raises the question of how to accurately define the ball's position when blur is present.

Traditionally, the convention for racket sports~\cite{sun2020,huang2019} has been to define the ball's position as the leading edge (or front) of the blur streak as shown in \Cref{fig:blur_example}.
However, this approach introduces ambiguities, such as determining which end of the streak is the front, and results in a non-symmetric representation that can complicate both detection and tracking.
A common workaround has been to provide the model with multiple sequential frames, allowing it to infer motion implicitly.
Although often dismissed as visual noise, motion blur inherently captures information about the ball’s velocity and direction of motion.
When properly leveraged, this visual cue becomes a powerful signal, enhancing both the robustness of ball tracking and the accuracy of trajectory prediction.

In this paper, we address the challenges of detecting fast-moving balls in high-speed sports like table tennis by explicitly modeling motion blur, a commonly ignored but informative visual cue.
Table tennis balls can reach speeds of up to 35m/s, making precise localization difficult, especially in blurred frames.
To address these challenges, we present the following contributions:
\begin{itemize}
\item \textbf{BlurBall}: A model that jointly predicts ball position and motion blur, achieving state-of-the-art performance.
\item \textbf{Blur-aware labeling}: A new annotation scheme that defines ball position at the blur center and encodes blur attributes, improving accuracy and resolving ambiguities.
\item \textbf{Table tennis ball dataset}: We introduce a new dataset for table tennis ball detection and tracking.
\end{itemize}

%% file: sec/2_related_work.tex
\section{Related Work}
\label{sec:related_work}

\textbf{Ball detection and tracking} have traditionally been approached using classical computer vision techniques, such as color filtering, background subtraction, and blob detection~\cite{archana2015, hung2018, tebbe2019, wong2023, tamaki2013}.
While these methods can be effective in controlled environments, they often struggle with variations in lighting, camera angles, and background clutter, requiring extensive fine-tuning to maintain performance.
Post-processing techniques can improve their robustness by enforcing physical constraints, such as filtering detections based on physically plausible trajectories~\cite{tamaki2013}.
However, these rule-based approaches remain sensitive to environmental changes and do not generalize well across different sports settings.

With the advancement of deep learning, neural network-based models have demonstrated greater robustness to dynamic environments and varying illumination conditions~\cite{zhao2017}.
Some approaches repurpose general object detectors, such as Detectron2~\cite{calandre2021a} and YOLOv4~\cite{kulkarni2022}, and fine-tune them on in-house sports datasets.
However, these models are not ideal for small, fast-moving objects like sports balls, which often appear as single instances per frame and can be heavily blurred.
To address these limitations, several specialized approaches have emerged that treat ball detection as a heatmap regression problem.
DeepBall~\cite{komorowski2019} was an early example of a fully convolutional network for generating ball-centered heatmaps, demonstrating effectiveness on soccer videos.
BallSeg~\cite{vanzandycke2019} is a modified version of ICNet~\cite{zhao2018icnet} applied to basketball videos.
TrackNet~\cite{huang2019}, developed for tennis, introduced Gaussian heatmaps to represent the probability distribution of ball locations.
Both previous methods introduced multiple consecutive input frames for implicit background subtraction and used the ball dynamics for filtering.
TrackNetV2~\cite{sun2020} extended this approach with a U-Net backbone~\cite{ronneberger2015} and adopted a multiple-input multiple-output (MIMO) design to jointly detect ball positions across several frames, improving robustness and inference speed.
Monotrack~\cite{liu2022b} further enhanced detection by adding residual connections and replacing the focal loss with a weighted combination of the Dice loss and the binary cross-entropy loss to better handle the extreme class imbalance caused by the ball's small size.
TrackNetV3~\cite{chen2024} addressed occlusion by incorporating a trajectory rectification module and an inpainting mechanism to recover missing detections.
The Widely Applicable Strong Baseline (WASB)~\cite{tarashima2023} switched from a U-Net to an HRNet~\cite{wang2021} backbone to achieve improved results.
Another key insight from it is that reducing the step size in the MIMO setup, effectively oversampling by shifting input windows by one frame instead of three, leads to notably better detection performance, albeit at the cost of slower inference.

\textbf{Attention Mechanisms.}
All the previously mentioned methods rely on CNNs, but transformers~\cite{vaswani2017} have proven highly effective for vision tasks such as segmentation and object detection, particularly with the introduction of Vision Transformers (ViTs)~\cite{vaswani2017, dosovitskiy2020}. 
While applying a full transformer architecture may be excessive for detecting a single small object, such as a ball, hybrid CNN-transformer models have emerged to capture temporal dependencies more effectively.
TrackNetV4~\cite{raj2024} builds on top of TrackNetV3 and introduces motion attention maps and a motion-aware fusion mechanism.
For table tennis, Li et al.~\cite{li2023} proposed a detector that integrates a transformer module with a CNN to leverage temporal context.
However, their dataset and code are not publicly available, which currently limits reproducibility.

As a lighter alternative, attention mechanisms within CNNs have gained traction for improving feature representation and detection accuracy.
While these mechanisms typically operate across channels, in the context of multi-frame inputs used in ball detectors, channel-wise attention can implicitly act as temporal attention, helping the model better capture motion cues and leverage temporal information.
We mostly focus on the following mechanisms.
The \ac{SE} block~\cite{hu2018} enhances performance by adaptively recalibrating channel-wise features. 
It generates a channel attention vector by applying global average pooling, followed by two fully connected layers with a ReLU activation, allowing the network to emphasize informative channels and suppress less relevant ones.
\ac{ECA}~\cite{wang2020} simplifies this process by removing the dimensionality reduction step and replacing the fully connected layers with a lightweight 1D convolution, enabling efficient local cross-channel interaction.
\ac{CA}~\cite{hou2021} extends channel attention by embedding spatial information through directional pooling along the horizontal and vertical axes. 
This allows the network to capture both content and location.
We propose integrating these attention mechanisms to improve ball detection performance.
Other attention mechanisms, such as \ac{CBAM} ~ \cite{woo2018}, \ac{GC} blocks~\cite{cao2020}, and non-local blocks~\cite{wang2018}, have also demonstrated improvements in object detection by combining channel and spatial attention or by modeling long-range dependencies.
However, due to their higher computational overhead and focus on broader contextual modeling, we leave their integration for future work and concentrate on methods most compatible with real-time, high-precision detection in high-speed sports scenarios.

\textbf{Racket sport datasets.}
The OpenTTGames dataset~\cite{voeikov2020a} is a publicly available resource for table tennis ball detection, collected under highly controlled conditions with high frame rates (120 fps), 1080p resolution, close-up views, and stable lighting.
However, such ideal conditions are rarely found in real-world broadcasts, where lower frame rates, lower resolutions, and inconsistent lighting make detection significantly more challenging.
Additionally, the dataset suffers from unbalanced camera color calibration, resulting in unnatural tones that hinder generalization to real-world footage.
Datasets for fast-moving objects (FMO) also include some table tennis footage~\cite{rozumnyi2017, kotera2019}, but with fewer than ten rallies, they are insufficient for training; the same holds for the synthetic VOT-FMO dataset~\cite{rozumnyi2021iccv}.
Ball-detection datasets exist for other sports, such as badminton~\cite{sun2020} and tennis~\cite{huang2019}.
Among them, the badminton shuttlecock is visually most similar to a table tennis ball, and models trained on badminton data~\cite{tarashima2023} can detect table tennis balls to some extent.
However, we observed that their detection accuracy on table tennis footage is significantly lower than the reported performance for badminton, highlighting the domain gap between the two tasks.
This highlights the necessity of a dedicated dataset for table tennis.

\textbf{Blur Estimation.}
Research on fast-moving objects has shown that motion blur can be explicitly modeled and exploited for trajectory reconstruction.
Rozumnyi et al.~\cite{rozumnyi2017} introduced the first dataset and tracker for FMOs, where objects appear as blurred streaks.
This line of work was extended with Tracking by Deblatting~\cite{kotera2019, rozumnyi2019, rozumnyi2021ijcv}, which combines blind deblurring and image matting to recover intra-frame trajectories from blur.
Further extensions enabled 3D reconstruction, using object size as a depth cue~\cite{zhang2023b, rozumnyi2020}.
Angular velocity estimation becomes also possible by finding the 3D rotation that best aligns successive blurred appearances~\cite{rozumnyi2020}.
More recently, FMODetect~\cite{rozumnyi2021iccv} proposed a learning-based detector trained on synthetic FMOs, achieving more efficient detection, trajectory estimation, and appearance recovery.  
However, most of these methods require labor-intensive labeling, such as high-speed camera annotations~\cite{kotera2019} or manual blur masks~\cite{rozumnyi2017}.
In contrast, for volleyball, Chao et al.~\cite{chao2024} attempted to implicitly exploit blur by constructing velocity heatmaps from consecutive frames, without explicitly modeling the underlying blur properties.
Unlike prior FMO methods, we assume blur follows a straight line, which simplifies dataset labeling and allows us to annotate a large number of diverse videos for training our model.

%% file: sec/3_method.tex
\section{Method}
\label{sec:method}
Traditional sports ball detectors are typically designed to output only the ball’s position, overlooking other valuable information available in a single frame, such as player positions and the field layout.
Among these, motion blur provides direct insight into the ball’s velocity.
Motion blur occurs when an object moves significantly during the camera’s exposure time, causing its appearance to stretch into a streak.
The magnitude of the blur is determined by the ball’s velocity and the exposure time: shorter exposures produce minimal blur, while longer exposures result in more pronounced streaks.

In this section, we first introduce our newly collected and labeled dataset for table tennis ball detection, which follows our proposed labeling convention that incorporates motion-blur information.
We then describe how this dataset is used to train BlurBall, our model that jointly estimates the ball’s position and its associated motion-blur properties.

\subsection{Dataset}\label{subsec:dataset}
\begin{figure}
\centering
\begin{subfigure}[b]{0.31\linewidth}
\centering
\includegraphics[width=\textwidth]{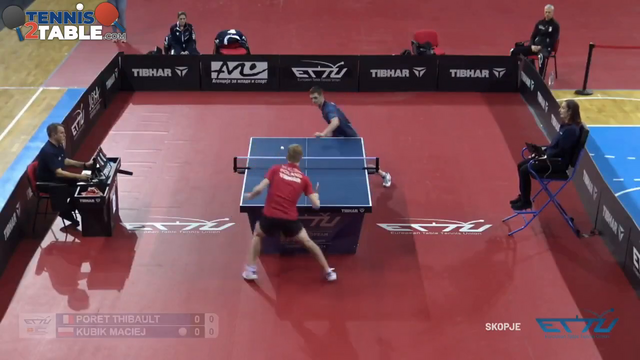}
\end{subfigure}
\hfill
\begin{subfigure}[b]{0.31\linewidth}
\centering
\includegraphics[width=\textwidth]{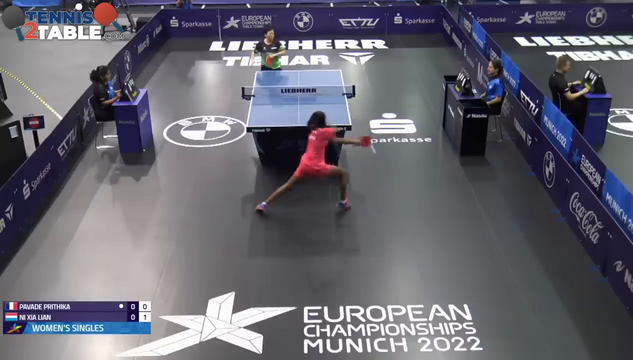}
\end{subfigure}
\hfill
\begin{subfigure}[b]{0.31\linewidth}
\centering
\includegraphics[width=\textwidth]{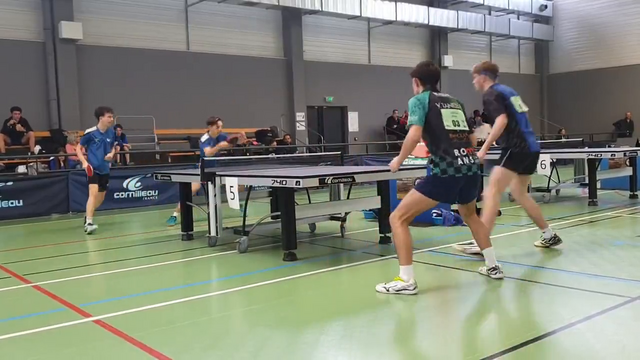}
\end{subfigure}

\begin{subfigure}[b]{0.31\linewidth}
    \centering
    \includegraphics[width=\textwidth]{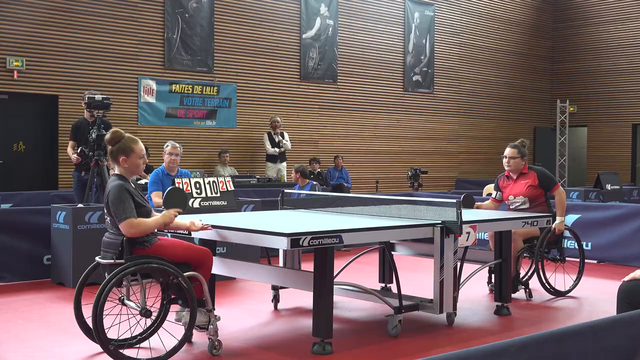}
\end{subfigure}
\hfill
\begin{subfigure}[b]{0.31\linewidth}
    \centering
    \includegraphics[width=\textwidth]{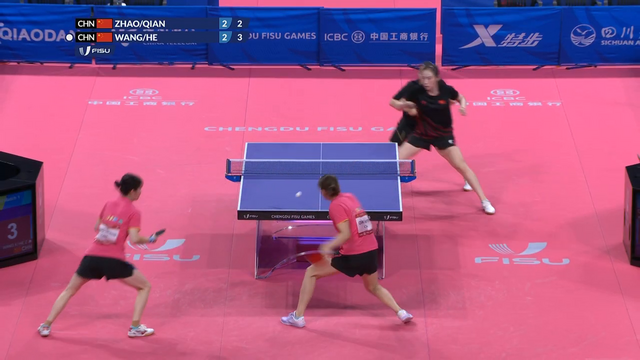}
\end{subfigure}
\hfill
\begin{subfigure}[b]{0.31\linewidth}
    \centering
    \includegraphics[width=\textwidth]{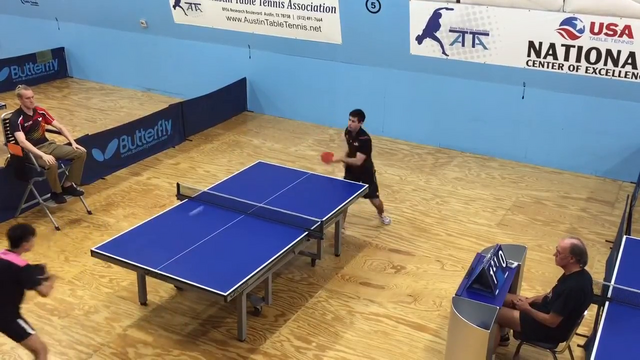}
\end{subfigure}

\begin{subfigure}[b]{0.31\linewidth}
    \centering
    \includegraphics[width=\textwidth]{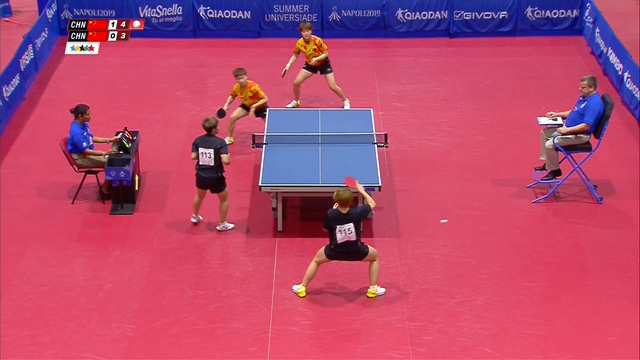}
\end{subfigure}
\hfill
\begin{subfigure}[b]{0.31\linewidth}
    \centering
    \includegraphics[width=\textwidth]{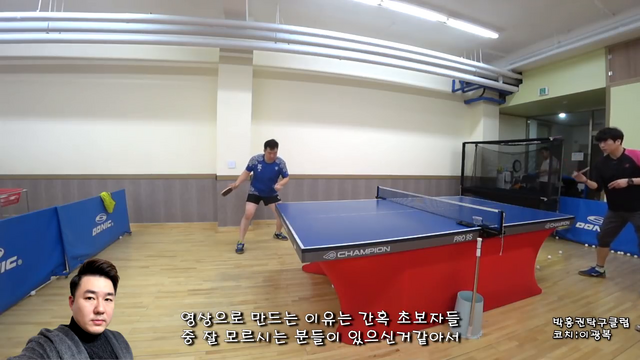}
\end{subfigure}
\hfill
\begin{subfigure}[b]{0.31\linewidth}
    \centering
    \includegraphics[width=\textwidth]{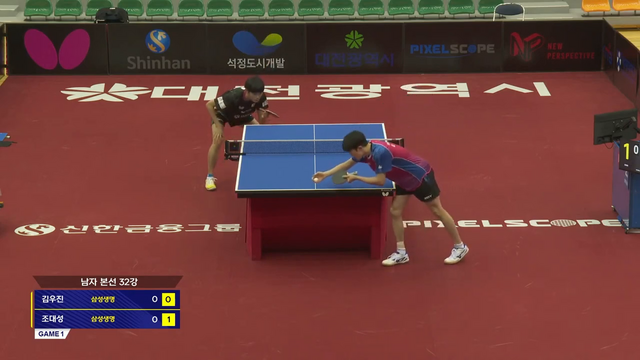}
\end{subfigure}

\begin{subfigure}[b]{0.31\linewidth}
    \centering
    \includegraphics[width=\textwidth]{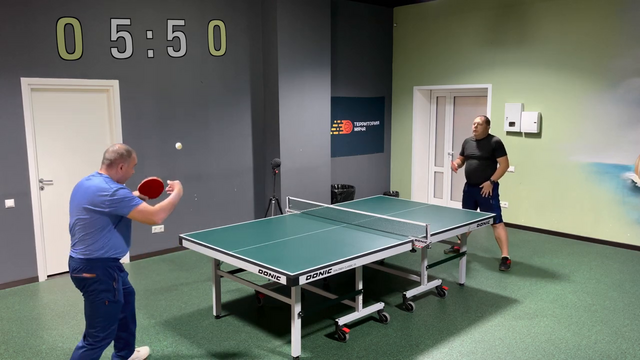}
\end{subfigure}
\hfill
\begin{subfigure}[b]{0.31\linewidth}
    \centering
    \includegraphics[width=\textwidth]{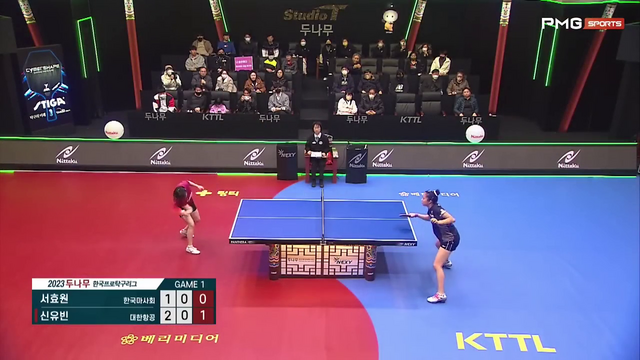}
\end{subfigure}
\hfill
\begin{subfigure}[b]{0.31\linewidth}
    \centering
    \includegraphics[width=\textwidth]{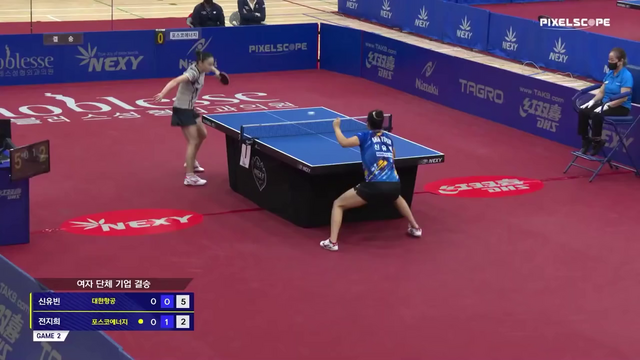}
\end{subfigure}

\caption{Example scenes from our dataset, showcasing a diverse range of contexts to ensure comprehensive coverage.}
\label{fig:dataset_grid}

\end{figure}

A comprehensive table tennis dataset was created from footage of 26 online recordings, spanning both amateur and professional games.
64,119 frames were gathered in total.
To enhance the dataset's diversity, we varied the point of view, table color schemes, lighting, and general environmental conditions as much as possible, as shown in \Cref{fig:dataset_grid}.
All videos were recorded from fixed, static camera positions.

Previous racket sport datasets~\cite{sun2020, huang2019} annotate the ball position at the leading edge of the motion blur streak.
In contrast, we propose a new labeling convention in which the ball position is defined as the midpoint of the blur streak.
Furthermore, we extend the labeling to include additional information about the blur streak, specifically its length and orientation, as shown in \Cref{fig:motion_blur_schematic}.
For each frame, we manually generated labels for the ball’s position, orientation, and blur length, represented as $[\bm{p_b},\theta,l]$, where $\bm{p_b}=(x_b,y_b)$ denotes the ball center.
These labels were obtained by manually drawing a line along the blur streak, like it is drawn in \Cref{fig:motion_blur_schematic}.
This is more time-effective than drawing the blur streak mask for each frame~\cite{rozumnyi2017} or labelling all frames from high-speed cameras~\cite{kotera2019}.
Approximating the blur streak as a straight line is justified, since the short exposure time ensures that the ball trajectory does not deviate significantly during image capture, except in specific rare cases as mentioned in \Cref{ssec:limitations}.
When motion blur is present, we define the angle $\theta$ and the half-length $l$ of the blur.
The two extremities of the blur streak are then calculated as follows:
\begin{equation}
\begin{aligned}
\mathbf{p}_1 =\; & \mathbf{p}_b + (l \cos \theta, l \sin \theta) \in \mathbb{R}^2, \\
\mathbf{p}_2 =\; & \mathbf{p}_b- (l \cos \theta, l \sin \theta) \in \mathbb{R}^2
\end{aligned}
\label{eq:line_blur}
\end{equation}
For comparison, we additionally include the traditional leading-edge labels i.e., the front of the ball blur streak.

\pgfkeys{/pgf/decoration/.cd,
distance/.initial=10pt
}

\pgfdeclaredecoration{add dim}{final}{
\state{final}{%
\pgfmathsetmacro{\dist}{5pt*\pgfkeysvalueof{/pgf/decoration/distance}/abs(\pgfkeysvalueof{/pgf/decoration/distance})}
\pgfpathmoveto{\pgfpoint{0pt}{0pt}}
\pgfpathlineto{\pgfpoint{0pt}{2*\dist}}
\pgfpathmoveto{\pgfpoint{\pgfdecoratedpathlength}{0pt}}
\pgfpathlineto{\pgfpoint{(\pgfdecoratedpathlength}{2*\dist}}
\pgfusepath{stroke}
\pgfsetdash{{0.1cm}{0.1cm}{0.1cm}{0.1cm}}{0cm}
\pgfsetarrowsstart{latex}
\pgfsetarrowsend{latex}
\pgfpathmoveto{\pgfpoint{0pt}{\dist}}
\pgfpathlineto{\pgfpoint{\pgfdecoratedpathlength}{\dist}}
\pgfusepath{stroke}
\pgfsetdash{}{0pt}
\pgfpathmoveto{\pgfpoint{0pt}{0pt}}
\pgfpathlineto{\pgfpoint{\pgfdecoratedpathlength}{0pt}}
}}

\tikzset{dim/.style args={#1,#2}{decoration={add dim,distance=#2},
decorate,
postaction={decorate,decoration={text along path,
raise=#2,
text align={align=center},
text={#1}}}}}

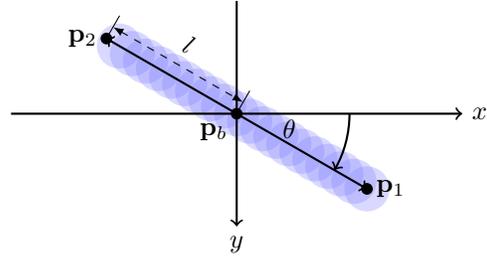
\begin{figure}
\centering

\begin{tikzpicture}
\coordinate (pb) at (0,0); 
\def\l{2} 

\coordinate (p1) at ({\l*cos(-30)}, {\l*sin(-30)});
\coordinate (p2) at ({-\l*cos(-30)}, {-\l*sin(-30)});

\foreach \t in {0,0.05,...,1} {
    \fill[blue!50, opacity=0.3] 
        ($ (p1)!\t!(p2) $) circle (0.3); 
}

\filldraw[black] (pb) circle (2pt) node[below left] {$\mathbf{p}_b$};

\draw[->, thick, black] (pb) -- (p1);
\filldraw[black] (p1) circle (2pt) node[right] {$\mathbf{p}_1$};

\draw[->, thick, black] (pb) -- (p2);
\filldraw[black] (p2) circle (2pt) node[left] {$\mathbf{p}_2$};

\draw[->, thick, black] (1.5,0) arc[start angle=0,end angle=-30,radius=1.5];
\node[black] at (0.7,-0.2) {$\theta$};

\draw[dim={$l$,10pt}] (p2) -- (pb);

\draw[thick,->] (-3,0) -- (3,0) node[right] {$x$};
\draw[thick,->] (0,1.5) -- (0,-1.5) node[below] {$y$};

\end{tikzpicture}
\caption{\textbf{Motion blur labeling schematic.} Conventional labels mark the front blur edge $\mathbf{p}_1$, which may be confused with $\mathbf{p}_2$ without motion context. We propose labeling the blur center $\mathbf{p}_b$ with its half-length $l$ and orientation $\theta$.}
\label{fig:motion_blur_schematic}
\end{figure}

\begin{table}
\centering
\resizebox{\columnwidth}{!}{%
\begin{tabular}{cccccc}
\hline
& Games & Clips & Frames & disp. [px] & Blur ratio \\
\hline
Train & 22 & 363 & 51423 & 18.7 $\pm$ 16.4 & 0.58\\
Test & 4 & 80 & 12696 & 20.1 $\pm$ 12.1& 0.77 \\
\hline
\end{tabular}
}
\caption{Table tennis dataset description. disp. is the mean ball displacement between frames. Blur ratio is the fraction of observations where the ball has motion blur.}
\label{tab:dataset}
\end{table}

\begin{figure}
\centering
\includegraphics[width=\linewidth]{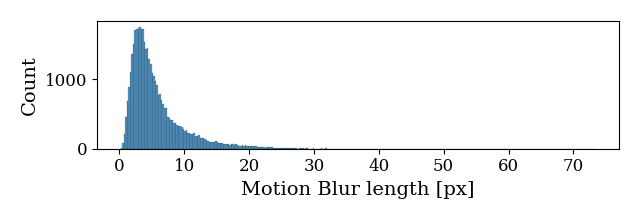}
\caption{Distribution of the half-lengths $l$ of the motion blur streak in our dataset. The maximum recorded value was 73 pixels.}
\label{fig:blur_dist}
\end{figure}

\Cref{tab:dataset} contains statistics that describe the generated dataset.
Because of the small size of the dataset, we limited ourselves to only a train and test set (80/20 ratio). 
We observe motion blur in 62\% of frames. 
The distribution of motion blur lengths is shown in \Cref{fig:blur_dist}, where we can see that the blur length is primarily concentrated around $l = 5$ pixels, although it can extend up to 73 pixels in some rare scenarios.
We split the dataset to maintain similar blur distribution in both test and train sets.

In addition to ball positions, we provide camera calibration data for each game, enabling 3D reconstruction of ball trajectories~\cite{gossard2025,kienzle2025}.
Specifically, we compute the camera’s extrinsic parameters, rotation $\bm{R}$ and translation $\bm{T}$, relative to the world frame, which is defined by the table tennis table.

\subsection{Ball Detector}
Among existing ball-detection and tracking methods, WASB~\cite{tarashima2023} demonstrated strong performance in tennis and badminton.
In WASB, an HRNet backbone generates heatmaps of likely ball locations, from which connected components above a threshold $\delta=0.5$ are selected as candidates.
A tracker then refines these candidates using tracklets from previous frames, ensuring temporal consistency and suppressing false positives.
We therefore adopt WASB as the basis of our model.
However, to effectively incorporate motion blur information, we introduced modifications to its training procedure.

Our key modification lies in the GT heatmap design.
In WASB, a binary map is generated using the following equation:
\begin{equation}
y_{\mathbf{p}}^{bin}= \begin{cases}
1 & \text{if } \left|\mathbf{p}-\mathbf{p}^{GT}\right| \leq d \\
0 & \text{otherwise}
\end{cases},
\end{equation}
where $\mathbf{p}^{GT}$ denotes the ground truth ball position, $y_{\mathbf{p}}^{bin}$ represents the value of the GT map at location $\mathbf{p}$, and $d$ is a distance threshold defining the radius of the disk.
In contrast, we redefine the ground truth (GT) heatmap to encompass the entire motion-blurred region of the ball as follows:
\begin{equation}
y_{\mathbf{p}}^{bin} =
\begin{cases}
1 & \text{if } \exists, \mathbf{p}^{\prime} \in [\mathbf{p}^{GT}_1, \mathbf{p}^{GT}_2] \text{ s.t. } |\mathbf{p} - \mathbf{p}^{\prime}| \leq d ,\\
0 & \text{otherwise}
\end{cases}
\end{equation}


In this formulation, all pixels within the blur streak, ranging from the detected start point $\bm{p_1}$ to the endpoint $\bm{p_2}$, are assigned a value of 1, effectively capturing the full extent of the motion blur of the ball.
This adaptation ensures that the model learns to localize not only the ball’s center but also the full extent of its motion blur.

To further refine the learning process, we adapted the Hard-to-Localize Sample Mining (HLSM) of WASB~\cite{tarashima2023} to our modified GT heatmap.
The original formulation of the GT heatmap is:
\begin{equation}
y_{\mathbf{p}}^{\text{real}} =
\begin{cases}
\min\left( C \exp\left( -\frac{|\mathbf{p} - \mathbf{p}^{GT}|^2}{d^2} \right), 1 \right) & \text{if } |\mathbf{p} - \mathbf{p}^{GT}| \leq d, \\
0 & \text{otherwise},
\end{cases}
\end{equation}
where $y_{\mathbf{p}}^{real}$ is the value of the real-valued GT map at $\mathbf{p}$ and $C$ is determined so that the non-zero minimum value is set to a predefined value $c_{min}$.
Similarly to the binary map, we extend the real-valued heatmap to encompass the blur with:
\begin{equation}
y_{\mathbf{p}}^{\text{real}} =
\begin{cases}
\max_{\mathbf{p}^{\prime} \in [\mathbf{p}_1^{GT}, \mathbf{p}_2^{GT}]}\;\min\left( C \exp\left( -\frac{|\mathbf{p} - \mathbf{p}^{\prime}|^2}{d^2} \right), 1 \right)
\\ \quad \text{if } \exists, \mathbf{p}^{\prime} \in [\mathbf{p}_1^{GT}, \mathbf{p}_2^{GT}] \text{ s.t. } |\mathbf{p} - \mathbf{p}^{\prime}| \leq d,
\\[0.8em] 
0 \quad \text{otherwise}.
\end{cases}
\end{equation}
%
%
The network was trained using the quality focal loss~\cite{li2021}, which helps the model focus on challenging regions where ball localization is difficult.
For inference, the center of the blur is calculated in a manner similar to WASB~\cite{tarashima2023}, where it is obtained as the weighted mean of the heatmap.
This ensures that the center is accurately located based on the intensity of the detected blur.

\begin{figure}
\centering
\begin{subfigure}[b]{0.45\linewidth}
\includegraphics[width=\textwidth]{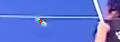}
\end{subfigure}
\hfill
\begin{subfigure}[b]{0.45\linewidth}
\includegraphics[width=\textwidth]{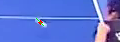}
\end{subfigure}

\begin{subfigure}[b]{0.45\linewidth}
    \includegraphics[width=\textwidth]{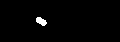}
\end{subfigure}
\hfill
\begin{subfigure}[b]{0.45\linewidth}
    \includegraphics[width=\textwidth]{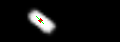}
\end{subfigure}
\caption{Left: Label and target heatmap. Right: Model prediction overlay with estimated ball center (red) and blur (green).}
\label{fig:ball_detector}

\end{figure}

To further improve the HRNet, we incorporate the attention mechanisms \ac{SE}, \ac{CA}, and \ac{ECA}, previously discussed in \Cref{sec:related_work}.
While \ac{SE} and \ac{ECA} improve channel-wise feature discrimination by capturing temporal cues in our multi-frame input with minimal computational overhead, \ac{CA} additionally encodes positional information, enabling more precise, spatially aware attention.
As shown in \Cref{sec:experiments}, \ac{SE} yields the most consistent performance improvement, and we choose this as our attention mechanism.
We refer to our model based on HRNet with \ac{SE} and trained for joint ball detection and blur estimation as \textbf{BlurBall}.
The training data and output predictions of BlurBall, including both ball localization and blur information, are visualized in \Cref{fig:ball_detector}.



\subsection{Estimating motion blur}
\label{ssec:ext_bad_dataset}
Once the trained BlurBall model produces a heatmap highlighting ball pixels, including those affected by motion blur, the next step is to process this heatmap to extract the blur parameters: orientation and length.
To isolate the blur region, we first apply a thresholding operation to the predicted heatmap.
By default, pixels with heatmap values over $\delta$ are considered part of the blur region.
To determine the direction of motion blur, we apply \ac{PCA} to the set of pixel coordinates within the extracted mask.
\ac{PCA} identifies the principal axis $\bm{B}$, which corresponds to the eigenvector associated with the largest eigenvalue.
This axis represents the primary direction of elongation in the blur streak and serves as an estimate of the ball’s motion direction during the camera’s exposure time.
The orientation angle $\theta$ of the blur is then computed as:
\begin{equation}
\theta = \arctan \left( \frac{B_y}{B_x} \right),
\end{equation}
which gives the blur direction relative to the horizontal axis.

To determine the blur length, we project all pixel coordinates $\mathbf{p}$ from the mask onto the principal axis.
The range of these projections represents the total extent of the blur along its dominant direction.
\begin{equation}
l = \frac{\max_{\mathbf{p} \in mask}(\bm{B}\cdot \mathbf{p}) - \min_{\mathbf{p} \in mask}(\bm{B}\cdot \mathbf{p})}{2}
\end{equation}
By extracting these two parameters, orientation ($\theta$) and length ($l$), we effectively recover the motion-blur characteristics, which provide direct insight into the ball's velocity.

In \cite{tarashima2023,sun2020}, the confidence score is calculated as the sum of the heatmap values within the detected blob region.
However, since our goal is to detect motion blur rather than the precise ball position, this approach tends to favor longer blur streaks, as the summed value naturally increases with streak length rather than signal strength.
To address this bias, BlurBall instead uses the mean value of the heatmap blob as the confidence score, making it more robust to variations in blur extent.

%% file: sec/4_experiments.tex
\section{EXPERIMENTS}
\label{sec:experiments}

\subsection{Blur estimation baseline}
\begin{figure}
\centering
\begin{subfigure}[b]{0.19\linewidth}
\centering
\includegraphics[width=\textwidth]{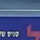}
\caption{}
\end{subfigure}
\hfill
\begin{subfigure}[b]{0.19\linewidth}
\centering
\includegraphics[width=\textwidth]{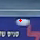}
\caption{}
\end{subfigure}
\hfill
\begin{subfigure}[b]{0.19\linewidth}
\centering
\includegraphics[width=\textwidth]{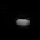}
\caption{}
\end{subfigure}
\hfill
\begin{subfigure}[b]{0.19\linewidth}
\centering
\includegraphics[width=\textwidth]{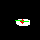}
\caption{}
\end{subfigure}
\hfill
\begin{subfigure}[b]{0.19\linewidth}
\centering
\includegraphics[width=\textwidth]{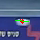}
\caption{}
\end{subfigure}
\caption{Baseline Method for Motion Blur Estimation. From left to right: (a) Frame $n-2$, (b) Frame $n$ with the estimated ball position, (c) Absolute difference between frames $n$ and $n-2$, (d) Binary mask obtained after thresholding, highlighting the estimated blur region, and (e) Frame $n$ with the detected blur overlaid.}
\label{fig:cv_blur}
\end{figure}

Existing ball detectors can be leveraged to extract the ball’s blur streak by incorporating additional standard computer vision techniques.
To establish a baseline for comparison, we implemented a method that detects motion blur using traditional image processing techniques, similar to \cite{rozumnyi2017}.
First, we define a small region of interest (ROI) centered around the detected ball position.
The input RGB image is then converted to grayscale, and we compute the frame difference to highlight regions of motion.
To suppress noise, a thresholding operation is applied, retaining only significant intensity changes.
The resulting binary mask might contain multiple connected components; we identify the one closest to the ROI center as the blur streak corresponding to the ball’s motion.
An example of this process is illustrated in \Cref{fig:cv_blur}.
Although this approach is not perfect, it can also be used to generate new labels using our convention for existing sports-ball datasets.
While this approach provides a straightforward way to estimate motion blur, it is inherently limited to static cameras and is less robust than BlurBall in dynamic environments.

\subsection{Training}
BlurBall was trained from scratch for 30 epochs using Adam.
We experimented with pretraining the network on the badminton dataset~\cite{sun2020}, which is the closest-looking sport to table tennis, but this did not yield any improvements.
The batch size was set to 8, and the ball detector takes as input images rescaled to 288 x 512.
We set $d$ to 2.5 and $c_{min}$ to 0.7, and start running HLSM at the beginning of the 20th epoch, as in the original WASB implementation.
Experiments were conducted on an NVIDIA GeForce RTX 2080 Ti.


\subsection{Ball Tracking}
\label{ssec:ball_tracking}

\begin{table*}
\centering
\resizebox{\linewidth}{!}{%
\begin{tabular}{l!{\vrule}cc|cc|cc|cc|cc|c|c}
\toprule
\multirow{2}{*}{\textbf{Model}} & \multicolumn{2}{c|}{\textbf{F1}} & \multicolumn{2}{c|}{\textbf{Acc}} & \multicolumn{2}{c|}{\textbf{AP}} & \multicolumn{2}{c|}{\textbf{Recall}} & \multicolumn{2}{c|}{\textbf{Precision}} & \textbf{\#Params} & \textbf{FPS} \\
& Front & Mid. & Front & Mid. & Front & Mid. & Front & Mid. & Front & Mid. & (M) & \\
\midrule
DeepBall~\cite{komorowski2019}       & 39.36   & 57.14   & 26.21   & 41.36   & 42.96 & 61.40   & 86.86 & 91.52   & 25.45 & 41.54 & 0.1 & 65 \\
\rowcolor[gray]{0.9} DeepBall-Large         & 52.40   & 71.72   & 36.93   & 56.66   & 55.42 & 75.03  & 89.82    & 94.12   & 36.99    & 57.94  & 1.1 & 51 \\
BallSeg~\cite{vanzandycke2019}        & 85.09   & 89.01   & 75.47   & 81.33   & 75.58 & 81.56   & 63.67 & 83.56   & 27.15 & 35.94   & 12.7 & 40 \\
\rowcolor[gray]{0.9} TrackNetV2~\cite{sun2020}    & 90.37   & 91.61   & 83.25   & 85.27   & 85.07 & 86.88   & 89.79 & 88.38   & 90.96 & 95.09   & 11.3 & 85 \\
ResTrackNetV2\footnotemark            & 87.03   & 91.81   & 78.11   & 85.61   & 75.53 & 86.58   & 86.76 & 89.02   & 87.90 & 94.78   & 1.2 & 112 \\
\rowcolor[gray]{0.9} Monotrack~\cite{liu2022b} & 93.60   & 94.97   & 88.55   & 90.89   & 87.97 & 91.26   & 93.13 & 94.93   & 94.07 & \textbf{96.36}  & 2.9 & \textbf{115} \\
TrackNetV3~\cite{chen2024}            & 93.74   & 95.93   & 88.35   & 92.28   & N.A.$^{\dagger}$ & N.A.$^{\dagger}$ & \textbf{99.64} & \textbf{99.66}   & 88.50 & 92.48   & 17.8 & 77 \\
\rowcolor[gray]{0.9} WASB (steps=3)~\cite{tarashima2023} & 94.23 & 95.77 & 89.62 & 92.26 & 91.82 & 94.66 & 93.89 & 95.23 & \textbf{94.57} & 96.31 & 1.5 & 95 \\
WASB (steps=1)~\cite{tarashima2023}   & \textbf{95.58} & \textbf{96.00} & \textbf{91.90} & \textbf{92.56} & \textbf{95.10} & \textbf{96.85} & 97.08 & 97.77 & 94.15 & 94.50 & 1.5 & 42 \\
\bottomrule
\end{tabular}%
}
\caption{Comparison of different models’ performance depending on the labeling convention. $^{\dagger}$AP is not available for TrackNetV3 because the rectification module doesn’t provide a confidence value.}
\label{tab:compare_labelling}
\end{table*}
\footnotetext{https://github.com/Chang-Chia-Chi/TrackNet-Badminton-Tracking-tensorflow2}

\begin{table*}
\centering
\resizebox{\linewidth}{!}{%
\begin{tabular}{l!{\vrule}cc|cc|cc|cc|cc|c|c}
\toprule
\multirow{2}{*}{\textbf{Model}} & \multicolumn{2}{c|}{\textbf{F1}} & \multicolumn{2}{c|}{\textbf{Acc}} & \multicolumn{2}{c|}{\textbf{AP}} & \multicolumn{2}{c|}{\textbf{MAE l} [px]} & \multicolumn{2}{c|}{\textbf{MAE $\theta$} [deg]} & \textbf{\#Params} & \textbf{FPS}  \\
& 1 step & 3 steps & 1 step & 3 steps & 1 step & 3 steps & 1 step & 3 steps & 1 step & 3 steps & (M) & for 3 step \\
\midrule
WASB~\cite{tarashima2023} (baseline) & 96.00 & 95.77 & 92.56 & 92.26 & 96.85 & 94.66 & 3.1$\pm$3.4$^{\dagger}$ & 3.0 $\pm$ 3.2$^{\dagger}$ & 13.5$\pm$23.2$^{\dagger}$ & 14.8 $\pm$ 24.6$^{\dagger}$ & 1.48 & 95 \\
\rowcolor[gray]{0.9} + Blur label & 96.28 & 95.54 & 93.05 & 91.81 & 97.59 & 95.40 & 1.5$\pm$1.2 & 1.4$\pm$1.2 & 6.5$\pm$18.2 & 7.2$\pm$20.1 & 1.48 & 95 \\
\quad + \ac{ECA} & 96.35 & 95.60 & 93.18 & 91.92 & 97.91 & 95.67 & 1.8$\pm$1.3 & 1.8$\pm$1.3 & \textbf{6.4$\pm$18.5} & 7.0$\pm$19.8 & 1.48 & 79 \\
\rowcolor[gray]{0.9} \quad + \ac{CA} & 96.50 & 95.76 & 93.48 & 92.23 & 97.24 & 94.82 & 1.4$\pm$1.2 & 1.4$\pm$1.2 & 6.5$\pm$18.6 & \textbf{6.7$\pm$18.8} & 1.50 & 72 \\
\quad + \ac{SE} $\rightarrow$ BlurBall (ours)& 96.52 & \textbf{96.16} & 93.47 & \textbf{92.89} & \textbf{98.23} & \textbf{96.72} & 1.5$\pm$1.2 & 1.6$\pm$1.2 & 6.5$\pm$18.9 & 6.9$\pm$19.5 & 1.49 & 79 \\
\midrule
\rowcolor[gray]{0.9} BlurBall with $\delta=0.7$ & \textbf{97.17} & 96.12& \textbf{94.75} & \textbf{92.89} & 97.34& 94.80 & \textbf{1.2$\pm$1.1} & \textbf{1.2$\pm$1.1} & 6.8$\pm$18.9 & 7.3$\pm$19.9 & 1.50 & 72 \\
\bottomrule
\end{tabular}%
}
\caption{Performance comparison starting from the WASB baseline, showing the effect of replacing the position-only heatmap with blur-aware heatmaps and evaluating different attention mechanisms. BlurBall, using SE attention, achieves the best overall performance. $^{\dagger}$WASB blur masks are obtained via background subtraction. Bold values indicate the best result in each column.}
\label{tab:ball_detector_eval}
\end{table*}

The effect of different labeling conventions on the performance of ball detectors was investigated.
\Cref{tab:compare_labelling} shows the performance of several sports ball detectors trained with two distinct labeling strategies: one where the ball position is labeled at the front edge of the blur streak (Front), and another where the ball’s position is defined at the midpoint of the blur (Mid.).
We used a fixed distance threshold of $\tau = 4\,\text{px}$ for the benchmark, as motivated in \cite{tarashima2023}.
WASB (1-step) achieves the best performance across all metrics except recall.
TrackNetV3 attains near-perfect recall, primarily due to its trajectory inpainting and rectification module.
Using the midpoint convention consistently improves detection performance across all models.
These results indicate that ball detectors perform better when the motion-blur streak is symmetric around the ball label, resulting in more accurate localization.
However, we did not notice any improvement in training speed with regard to the labeling convention used.
Given these findings, we recommend adopting this midpoint labeling convention for ball detectors, particularly in fast-paced sports where motion blur is common.
This approach enhances the robustness and accuracy of detection systems in such dynamic environments.

As shown in \Cref{tab:ball_detector_eval}, incorporating blur information into WASB with a 3-step setting slightly decreases the F1 score and accuracy, yet it improves the AP.
In contrast, for the 1-step setting, adding blur information consistently enhances all metrics, suggesting that using the blur from adjacent frames helps infer the ball’s likely location in the center frame more accurately.

Introducing attention mechanisms improves performance across all metrics compared to the blur-augmented baseline.
Among them, the \ac{SE} block yields the most consistent improvements.
Its ability to globally recalibrate channel-wise feature responses allows it to emphasize both motion and appearance features.
In comparison, \ac{CA} incorporates positional information through coordinate attention, which can be beneficial for larger or spatially consistent objects.
However, in our case, where the target is small, fast-moving, and lacks stable spatial patterns, \ac{CA}’s added complexity, including per-axis pooling and normalization, may introduce noise, particularly in the 3-step setup.
\ac{ECA} offers a lighter alternative by using local 1D convolutions without dimensionality reduction, but this comes at the cost of modeling only short-range dependencies.
In practice, it lacks the global context modeling required for robust motion-blur interpretation.
While \ac{CA} and \ac{SE} achieve similar performance in the 1-step setting, \ac{SE} is both slightly faster and more robust across settings.
For these reasons, we select \ac{SE} as the attention mechanism for BlurBall.

\begin{figure}
\centering
\includegraphics[width=0.9\linewidth]{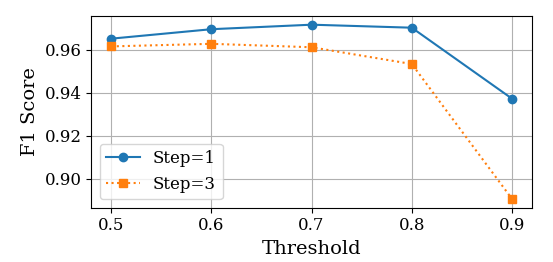}
\caption{F1 score for BlurBall with different threshold values $\delta$ on the test set.}
\label{fig:f1_score}
\end{figure}

Although WASB achieves the highest accuracy in the 1-step setting, its precision drops compared to the 3-step version.
This suggests that the model is overconfident in the central frame predictions and that the default threshold of $\delta=0.5$ is suboptimal.
As shown in \Cref{fig:f1_score}, the threshold has little influence for 3-step inference, but for 1-step the optimal value shifts to $\delta=0.7$.
With this adjustment, BlurBall obtains the best F1-score and accuracy (\Cref{tab:ball_detector_eval}), though at the cost of a slight decrease in AP and blur estimation quality.

%


\subsection{Blur Estimation}
We evaluate and compare the accuracy of BlurBall and WASB in \Cref{tab:ball_detector_eval}.
The motion blur estimation performance is assessed using the \ac{MAE} for both blur length $l$ and blur angle $\theta$.
For small blur lengths, particularly when $l \leq 3$, the estimation of the blur angle becomes highly sensitive to noise, rendering the measurements unreliable.
To ensure robust evaluation, the \ac{MAE} for the blur angle is computed only for predictions with an estimated blur length exceeding 3 pixels.
Under this constraint, BlurBall proves significantly more accurate, achieving nearly twice the precision of WASB combined with the baseline blur estimation.
Among the attention mechanisms tested, \ac{SE} and \ac{CA} improve blur estimation, whereas \ac{ECA} does not offer any notable benefit.

\subsection{Ball Trajectory Prediction}
\label{ssec:ball_traj_pred}
In this subsection, we demonstrate that motion blur provides valuable cues for predicting the ball’s trajectory.
The ball’s 3D trajectory can be reasonably approximated using a second-degree polynomial.
This assumes that the forces acting on the ball (gravity, drag, and Magnus force) remain relatively constant while the ball is airborne.
Consequently, the 2D projection of the trajectory in the image plane can also be modeled as a second-degree polynomial.
We represent the ball’s image coordinates over time as $(P_x(t), P_y(t))$, where $P_x$ and $P_y$ are quadratic polynomials with respect to time.

\begin{figure}
\centering
\includegraphics[width=\linewidth]{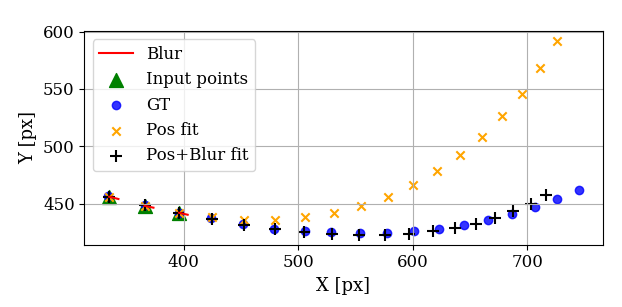}
\caption{Trajectory prediction comparison using position-only vs. position+blur fitting. Polynomials are fitted using the first three observations to predict the ball’s trajectory. Predictions are compared to the ground truth (GT) positions. Red lines indicate observed motion blur.}
\label{fig:traj_pred}
\end{figure}

In this context, we show that incorporating motion-blur information improves trajectory prediction during flight.
Motion blur is directly linked to the ball’s speed and camera exposure time $t_{\text{exp}}$, allowing us to relate the observed blur length $l$ and angle $\theta$ to the derivative of the trajectory as follows:
\begin{equation}
\begin{aligned}
\dot{P}_x (t) &= \frac{l \cos(\theta)}{t_{\text{exp}}}, \quad
\dot{P}_y (t) = \frac{l \sin(\theta)}{t_{\text{exp}}}.
\end{aligned}
\end{equation}
For evaluation, we use BlurBall to infer both the ball position and motion blur.
To predict the ball’s trajectory, we fit a second-degree polynomial using only the first 3 frames of each sequence, the minimum required to estimate a quadratic curve.
Trajectory fitting based solely on position is performed using a standard least-squares method.
For position + blur fitting, we minimize the following cost functions using the Nelder-Mead algorithm to obtain the polynomial coefficients and $t_{\text{exp}}$:
\begin{align}
J_x &= \frac{1}{3} \sum_{k=0}^{2} \left| P_x (t_k) - \hat{x}_k \right|^2
+ 0.2 \left| \dot{P}_x (t_k) - \frac{l_k \cos(\theta_k)}{t_{\text{exp}}} \right|^2 \\
J_y &= \frac{1}{3} \sum_{k=0}^{2} \left| P_y(t_k) - \hat{y}_k \right|^2
+ 0.2 \left| \dot{P}_y (t_k) - \frac{l_k \sin(\theta_k)}{t_{\text{exp}}} \right|^2.
\end{align}
Here, $(\hat{x}_k, \hat{y}_k)$ are the detected positions, $l_k$ is the estimated blur length, and $\theta_k$ is the blur angle for each frame $k$.
We weight the blur loss with a coefficient of 0.2.
After fitting the polynomial, we compare the predicted trajectory to the ground truth positions for the remainder of the ball’s flight.
This evaluation is performed on 95 manually segmented trajectories from the test set.
An example of the predicted trajectory is shown in \Cref{fig:traj_pred}.

We obtain a \ac{MAE} of $84.4\pm136.6$ pixels for the trajectory fitting using position only, and a significantly lower \ac{MAE} of $53.0\pm87.1$ pixels when incorporating both position and blur information.
The notable reduction in both mean and variance of the prediction error demonstrates that motion blur encodes meaningful velocity information.
Incorporating these cues yields more accurate and consistent trajectory estimates, especially in fast-motion scenarios or when only a few frames are available.

\subsection{Limitations}
\label{ssec:limitations}

\begin{figure}
\centering
\begin{subfigure}[b]{0.23\linewidth}
\includegraphics[width=\textwidth]{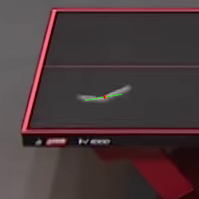}
\caption{Input 1}
\end{subfigure}
\hfill
\begin{subfigure}[b]{0.23\linewidth}
\includegraphics[width=\textwidth]{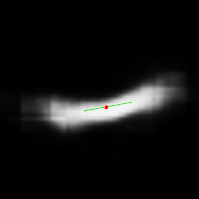}
\caption{HM 1}
\end{subfigure}
\hfill
\begin{subfigure}[b]{0.23\linewidth}
\includegraphics[width=\textwidth]{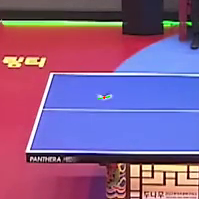}
\caption{Input 2}
\end{subfigure}
\hfill
\begin{subfigure}[b]{0.23\linewidth}
\includegraphics[width=\textwidth]{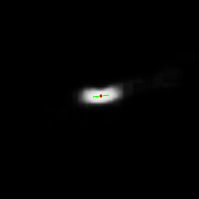}
\caption{HM 2}
\end{subfigure}
\caption{Frames and corresponding inferred heatmaps where the bounce is exactly captured by the camera.}
\label{fig:bounce_blur}
\end{figure}

BlurBall has certain limitations.
One key assumption in our approach is that motion blur appears as a single linear streak.
However, in rare cases, such as when the exact moment of a ball bounce is captured, this assumption no longer holds, as illustrated in \Cref{fig:bounce_blur}.
The ball is still detected, but the estimated position is inaccurate because the detector uses the center of a non-convex blob, which is distorted by the bounce.
Still, the output heatmap seems to have captured the change in direction to some degree, as shown by the bend in the heatmap blob.
Also, although the model can work for videos with non-static cameras, we observed that it is more prone to misdetections.
This introduces potential challenges in accurately estimating the ball’s motion in such cases.

While our current model achieves state-of-the-art ball detection for table tennis, some failure cases remain and highlight areas for future improvement.
A common source of false positives involves white objects being misidentified as the ball.
This typically includes shoes, hands, or logos on uniforms that resemble the ball in size, shape, or motion.
Examples of such cases are shown in the supplementary material.
To reduce false positives, future work could include additional context from the entire scene to help distinguish the ball from similar-looking objects.

%% file: sec/5_conclusion.tex
\section{CONCLUSION}
\label{sec:conclusion}

In this paper, we introduced a novel labeling convention for sports ball detectors that explicitly incorporates motion blur, significantly improving detection performance across all models tested on our newly introduced table tennis dataset.
We also presented BlurBall, a detector that jointly estimates ball position and motion blur, providing valuable velocity cues and enabling more accurate trajectory prediction.
Our model achieves competitive performance by integrating attention mechanisms, which benefit from multi-frame inputs.
Our approach enables more precise and robust tracking, with strong potential for broader applications in high-speed sports analytics and real-world deployments.

%% file: sec/appendix.tex
{
\newpage
   \twocolumn[
    \centering
    \Large
    \textbf{BlurBall: a ball detector with blur estimation}\\
    \vspace{0.5em}Supplementary Material \\
    \vspace{1.0em}
   ] 
}

\section*{Dataset}
The dataset was generated from publicly available online videos under a Creative Commons license to ensure compliance with copyright regulations.
We deliberately covered a wide range of scenarios, including different playing conditions, camera angles, and lighting variations, to improve model robustness.
Orange table tennis balls, although officially approved for competition, are not included. 
This omission is unlikely to affect performance, as white balls are overwhelmingly preferred and almost exclusively used in professional matches.

The dataset is publicly available at: \url{https://cogsys-tuebingen.github.io/blurball/}.
Each video is accompanied by a CSV file with the following structure:

\begin{table}[h]
    \centering
    \begin{tabular}{|c|c|c|c|c|c|}
    \hline
         Frame & Visibility & X & Y & $\theta$ & $l$ \\
         \hline
        000049 & 1 & 581.62 & 295.26 & -152.5 & 2.8 \\
        000050 & 1 & 572.98 & 292.86 & 171.8  & 2.1 \\
         \hline
    \end{tabular}
    \caption{CSV description. Each row contains the ball position ($X, Y$), blur orientation ($\theta$), and half-length ($l$).}
    \label{tab:csv_desc}
\end{table}

Angles are given in degrees and follow the convention illustrated in \Cref{fig:motion_blur_schematic}.

\subsection*{Camera calibration}
\begin{figure}
    \centering
    \includegraphics[width=0.8\linewidth]{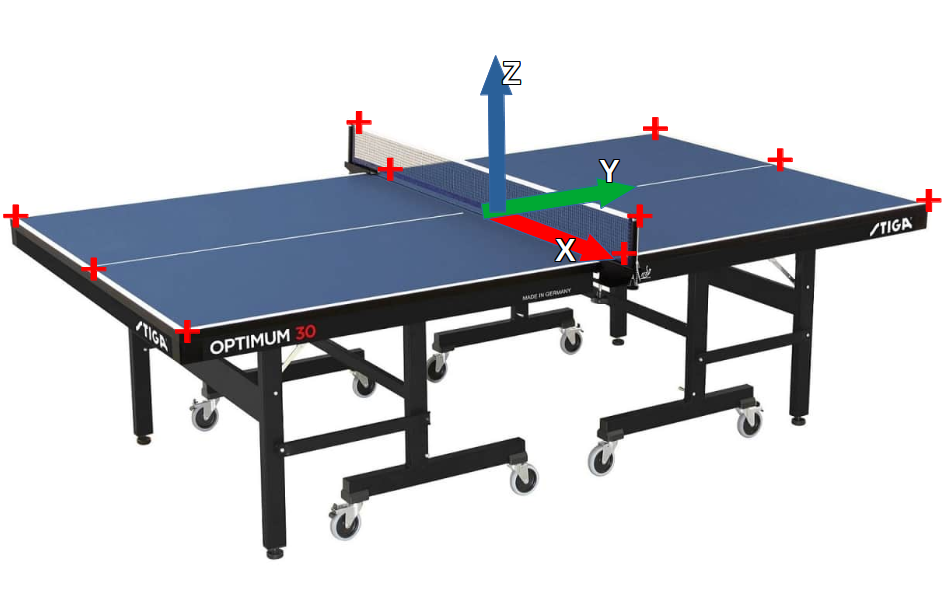}
    \caption{Annotated table keypoints used for camera calibration and definition of the world frame.}
    \label{fig:table_kp}
\end{figure}

We provide camera calibration for each table tennis match in the dataset.
Specifically, we include the focal length $f$ and the camera extrinsics: rotation vector $\bm{r}$ and translation vector $\bm{T}$.
The world frame is defined by the table geometry, with the table’s length aligned to the $Y$-axis and the surface normal aligned to the $Z$-axis, as shown in \Cref{fig:table_kp}.

This choice is motivated by the precise, known geometry of the table, which makes it a reliable calibration target.
We manually annotate keypoints such as the four table corners, the midline–backline intersection, and the intersection between the net and the table’s side edge (\Cref{fig:table_kp}). 
The camera pose is then estimated by minimizing the reprojection error via a standard PnP optimization.

Due to the limited and near-coplanar set of keypoints, distortion coefficients $d_d$ and the optical center cannot be reliably estimated. 
We therefore assume an ideal pinhole model, with intrinsic parameters reduced to the focal length $f$.

\section*{Model training}
We followed the training setup of WASB~\cite{tarashima2023} for consistency across models.
Each model was trained for 30 epochs with an appropriate loss function and optimizer:

\begin{itemize}
    \item \textbf{DeepBall}~\cite{komorowski2019} and \textbf{DeepBall-large}: BCE loss, Adam optimizer, learning rate $3 \times 10^{-4}$.  
    \item \textbf{BallSeg}~\cite{vanzandycke2019}: focal loss ($\gamma = 2$), Adam optimizer, learning rate $3 \times 10^{-4}$.  
    \item \textbf{TrackNetV2}~\cite{sun2020} and \textbf{ResTrackNetV2}: focal loss, AdaDelta optimizer, learning rate $1.0$.  
    \item \textbf{MonoTrack}~\cite{liu2022b}: Combo loss, AdaDelta optimizer, learning rate $1.0$.  
    \item \textbf{WASB}~\cite{tarashima2023}: quality focal loss, Adam optimizer, learning rate $3 \times 10^{-4}$.  
\end{itemize}

This ensures fair comparison, with each model optimized using strategies suited to its architecture.

\section*{BlurBall: further insights}

\subsection*{Impact of the threshold value}
\begin{figure}
    \centering
    \includegraphics[width=\linewidth]{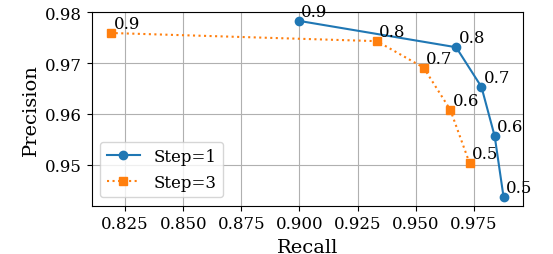}
    \caption{PR curves for BlurBall at different confidence thresholds $\delta$.}
    \label{fig:pr_curve}
\end{figure}

To evaluate the influence of the detection threshold in our BlurBall model, we plot the Precision-Recall (PR) curve in \Cref{fig:pr_curve}.
Overall, the 1-step variant achieves a better balance than the 3-step variant.
However, for identical threshold values, the 1-step detector consistently yields higher recall but lower precision, indicating that its middle-frame predictions tend to be overconfident.
This trade-off motivates the choice of a threshold value of $\delta = 0.7$, which we adopt for subsequent tracking experiments in \Cref{ssec:ball_tracking}.

\subsection*{Blur prediction}
\begin{figure}
    \centering
    \includegraphics[width=\linewidth]{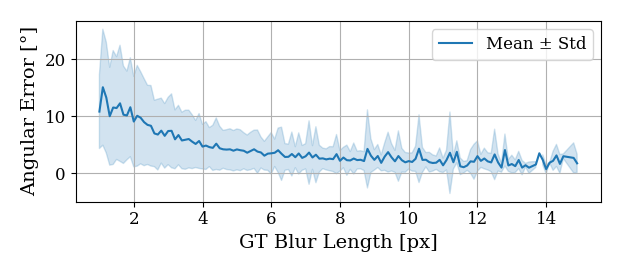}
    \caption{Angular estimation error as a function of ground-truth blur length.}
    \label{fig:blur_dist_error}
\end{figure}

\Cref{fig:blur_dist_error} shows the relationship between blur length and angle estimation error.
As expected, longer blur streaks yield more accurate angle estimation. 
For $l > 3$, the angular error remains below $10^\circ$, validating its use in downstream tasks such as trajectory prediction (\Cref{ssec:ball_traj_pred}).

\subsection*{Failure cases}
\begin{figure}
    \centering
    \begin{subfigure}[b]{0.31\linewidth}
        \includegraphics[width=\textwidth]{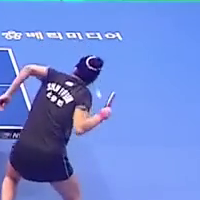}
        \caption{FN1}
    \end{subfigure}
    \hfill
    \begin{subfigure}[b]{0.31\linewidth}
        \includegraphics[width=\textwidth]{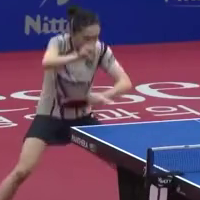}
        \caption{FN2}
    \end{subfigure}
    \hfill
    \begin{subfigure}[b]{0.31\linewidth}
        \includegraphics[width=\textwidth]{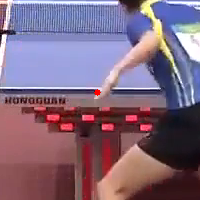}
        \caption{FP1}
    \end{subfigure}
    \caption{Examples of BlurBall failures. FN1 and FN2: missed detections. FP1: false positive on a hand.}
    \label{fig:blurball_failure}
\end{figure}

\Cref{fig:blurball_failure} shows typical failure cases.
In FN1, the ball briefly appears between two occluded frames, leading to a missed detection.
In FN2, the ball overlaps with a moving player’s body, a scenario difficult even for human observers.
In FP1, a hand is incorrectly detected as the ball.

\section*{Trajectory prediction benchmark}
\begin{figure}
    \centering
    \includegraphics[width=\linewidth]{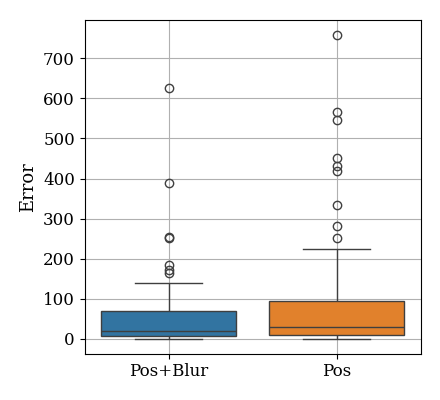}
    \caption{Full box plot of trajectory prediction errors for \texttt{Pos} and \texttt{Pos+Blur}, including outliers.}
    \label{fig:traj_pred_box_full}
\end{figure}

\begin{figure}
    \centering
    \includegraphics[width=\linewidth]{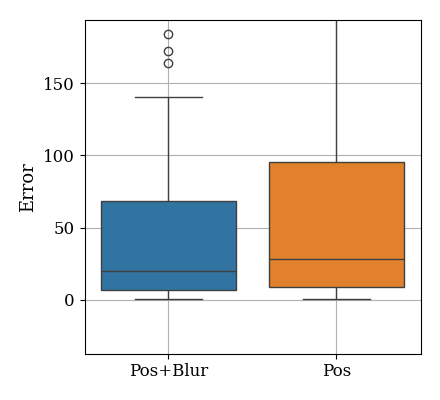}
    \caption{Zoomed-in box plot of trajectory prediction errors.}
    \label{fig:traj_pred_box_zoom}
\end{figure}

We benchmark trajectory prediction using position-only (\texttt{Pos}) versus position+blur (\texttt{Pos+Blur}) fitting. 
\Cref{fig:traj_pred_box_full} and \Cref{fig:traj_pred_box_zoom} report full and zoomed-in box plots.

As shown in \Cref{fig:traj_pred_box_zoom}, \texttt{Pos+Blur} achieves a lower median error (19.9\,px vs.\ 28.4\,px) and a narrower interquartile range, indicating greater consistency.
The full distribution (\Cref{fig:traj_pred_box_full}) also reveals fewer extreme outliers and a shorter upper whisker (140.2\,px vs.\ 224.6\,px), confirming improved robustness in difficult cases.
Overall, incorporating blur leads to more accurate and stable trajectory predictions, especially under sparse or noisy observations.